\documentclass[10pt,twocolumn,letterpaper]{article}

\usepackage{cvpr}
\usepackage{times}
\usepackage{epsfig}
\usepackage{graphicx}
\usepackage{amsmath}
\usepackage{amssymb}
\usepackage{mathptmx}
\usepackage{caption}
\usepackage{subcaption}
\usepackage{comment}

\usepackage{url}            
\usepackage{booktabs}       
\usepackage{amsfonts}       
\usepackage{nicefrac}       
\usepackage{microtype}      
\usepackage{algorithm}
\usepackage{algorithmic}

\usepackage{helvet} 
\usepackage{courier}  
\usepackage{multirow}
\usepackage{amsmath,amssymb}
\usepackage{makecell}

\usepackage[pagebackref=true,breaklinks=true,letterpaper=true,colorlinks,bookmarks=false]{hyperref}

\cvprfinalcopy 

\ifcvprfinal\fi
\begin{document}

\title{3DPVNet: Patch-level 3D Hough Voting Network for 6D Pose Estimation}

\newcommand*\samethanks[1][\value{footnote}]{\footnotemark[#1]}

\author{
	Yuanpeng Liu$^{1}$,
	Jun Zhou$^{1}$,
	Yuqi Zhang$^{1}$,
	Chao Ding$^{1}$,
	Jun Wang$^{1}$\\
$^{1}$Nanjing University of Aeronautics and Astronautics \\
}

\maketitle

\begin{abstract}

In this paper, we focus on estimating the 6D pose of objects in point clouds.
Although the topic has been widely studied, pose estimation in point clouds remains a challenging problem due to the noise and occlusion.
To address the problem, a novel 3DPVNet is presented in this work, which utilizes 3D local patches to vote for the object 6D poses.
3DPVNet is comprised of three modules. 
In particular, a Patch Unification (\textbf{PU}) module is first introduced to normalize the input patch, and also create a standard local coordinate frame on it to generate a reliable vote.
We then devise a Weight-guided Neighboring Feature Fusion (\textbf{WNFF}) module in the network, which fuses the neighboring features to yield a semi-global feature for the center patch. WNFF module mines the neighboring information of a local patch, such that the representation capability to local geometric characteristics is significantly enhanced, making the method robust to a certain level of noise.
Moreover, we present a Patch-level Voting (\textbf{PV}) module to regress transformations and generates pose votes. After the aggregation of all votes from patches and a refinement step, the final pose of the object can be obtained.
Compared to recent voting-based methods, 3DPVNet is patch-level, and directly carried out on point clouds.
Therefore, 3DPVNet achieves less computation than point/pixel-level voting scheme, and has robustness to partial data.
Experiments on several datasets demonstrate that 3DPVNet achieves the state-of-the-art performance, and is also robust against noise and occlusions.

\end{abstract}

\section{Introduction}

\label{sec:introduction}
\noindent Detecting objects and estimating their 3D translations and 3D rotations with respect to the camera coordinate frame, known as the 6D pose, are widely studied topics in various fields, such as robotic manipulation and scene understanding.
Although RGB-only pose estimation methods have shown impressive performances, point clouds can provide additional geometric information, making the estimation results more accurate.
Recent developments on 3D sensors, such as depth cameras, have made it convenient for researchers to collect 3D data, facilitating the study of pose estimation on point clouds.
However, the point cloud acquired by these devices usually contains a certain level of noise, and the objects of interest are unevenly partial, sometimes occluded by barriers, making the pose estimation task fair challenging.

Recently, voting-based methods have shown significant effectiveness for the task, while most of them utilize deep learning methods to learn features and vote for object pose. 
However, these methods are mainly carried out on 2D or 2.5D images, such as PVNet~\cite{peng2019pvnet}.
DenseFusion\cite{wang2019densefusion} fuses the features from the image and point cloud to obtain a representative feature.
VoteNet\cite{qi2019deep} simply votes for the object location, which cannot be directly applied to pose estimation task.
Consequently, voting methods in point clouds for pose estimation based on deep learning are rarely studied.
Actually, voting for object pose in point clouds suffers the following challenges.
First, for a voting unit (point/patch), the coordinate frame is the based on the global coordinate frame (\textit{GCF}), which is unstable as the viewpoint to the scene changing.
Second, the feature learned from a single unit may not be representative, since the contextual and geometric cues in a local region are hard to be captured by one unit.
Third, the magnitude of the unit significantly affects the estimation accuracy and efficiency.
Existing methods~\cite{peng2019pvnet, wang2019densefusion, he2020pvn3d} employ point/pixel-wise voting, which account for a large amount of computation redundancy.

In this paper, we propose 3DPVNet, a patch-level Hough voting method based on deep pose regression for 6DoF pose estimation of known objects from point clouds.
Our method is inspired by an intuition that one can roughly infer the object pose through observing only a 3D local patch on its surface. The intuition implies that the local patch contains necessary information which can be used for inferring the object 6D pose. 
Based on the intuition, we present to directly regress 6D poses using 3D patches. 
The original input is RGB-D images.
We adopt a semantic segmentation network to extract objects of interest from RGB images, which are then convert into point cloud using depth maps.

3DPVNet contains three modules to address the mentioned challenges. 
The first is the Patch Unification (PU) module, which aims to create a normalized patch for the input of the network and a standard local coordinate frame on the patch for stably voting.
More specifically, we build a Patch Normalization branch in the module to achieve the normalized input. In order to obtain a reliable pose vote, which stands for the transformation from the object coordinate frame (\textit{OCF}) to the \textit{GCF}, we construct a stable local patch coordinate frame (\textit{LPCF}) through the Patch Standardization branch. The standardization process computes the transformation from \textit{GCF} to the \textit{LPCF}.
Since there are not enough contextual and geometric cues on a local patch to avoid the similar structures, we propose using the neighboring patches to supplement the representativeness. Specifically, we utilize a backbone to extract features from patches, and devise a Weight-guided Neighboring Feature Fusion (WNFF) module in the network to fuse all features from the input patch and neighboring patches according to their individual weights. The weight is allocated adaptively, which distinguishes the contributions of different features commendably. The module can produce a semi-global feature that contains additional geometric characteristics, making the feature of a local patch more representative.
Finally, we design a Patch-level Voting (PV) module to regress the transformation vectors using the output semi-global features. The transformation vector represents the transformation from the \textit{LPCF} to the \textit{OCF}, which is combined with the transformation from standardization branch to generate a pose votes.
All votes are aggregated through a pose clustering step to yield a relatively rough 6D pose, which are then refined to obtain the final 6D pose. 

Compared to previous works, our method has two key characteristics. 
First, our method is patch-level, and involves a set of definitions and transformations of coordinate frames for 6D pose estimation in point clouds, which are rarely discussed. Moreover, the patch-level voting scheme is less computational than point/pixel-level voting methods, without any loss of accuracy of pose estimation.
Second, we propose fusing the features of the input and neighboring patches according to their individual weights, which significantly enhances the capture of geometric cues

\section{Related work}
\label{sec:related_work}
Over the years, various voting-based methods presented to deal with 6D pose estimation problem from the 3D point cloud. We briefly review these methods related to our work from the following aspects.

\noindent \textbf{Voting-based Methods.} 
A classical work is \cite{drost2010model}, which constructed the point pair features (PPFs) on both 3D models and scenes. A matching process is then implemented based on hash tables. Each successful match yields a transformation to vote for the final pose. However, the method shows an expensive time consumption. 
The developed variations~\cite{choi2012voting, salas2013slam++, birdal2015point, hinterstoisser2016going, li2018fast} show a significant improvement in terms of accuracy and cost.
In particular, \cite{hinterstoisser2016going} improved the robustness of PPF to the sensor noise and background clutter by applying an efficient sampling strategy, which achieves satisfactory performance on several objects from challenging benchmarks.
Recently, voting-based pose estimation methods gain more attention. \cite{jafari2018ipose} constructs an encoder-decoder system for mapping pixels to 3D positions after segmentation module. 
\cite{oberweger2018making} uses image patches to predict 2D heatmap, which serves as 2D projections of 3D points.
PVNet~\cite{peng2019pvnet} predicts vectors pointing to 2D keypoints, forming dense correspondences between 2D and 3D data. 
Moreover, PVN3D~\cite{he2020pvn3d} extends the method to regress 3D keypoints directly by adding extra depth information. 
Hu et al.~\cite{hu2019segmentation} proposed a two-stage framework that establish 3D-to-2D correspondences with segmentation patches vote to 2D keypoints, then adopted the RANSAC algorithm to estimate pose. 
\cite{hu2020single} integrates PnP algorithm into the network, which hence directly regresses pose from correspondences. 

\noindent \textbf{Deep Learning on Point Clouds.} 
There are series of deep learning works on point clouds have been presented over these years, such as PointNet~\cite{qi2017pointnet}, PointNet++~\cite{qi2017pointnet++}, PointCNN~\cite{li2018pointcnn}, PointSIFT~\cite{jiang2018pointsift}, Splatnet~\cite{su2018splatnet}, Dynamic Graph CNN~\cite{wang2019dynamic}, etc.
Based on these deep learning architectures, many researchers seek to exploit creative applications on point clouds, including model classification~\cite{cao20173d, roynard2018classification}, semantic segmentation~\cite{tchapmi2017segcloud, engelmann2017exploring, graham20183d}, and object detection~\cite{zhou2018voxelnet, qi2019deep}. 
\cite{qi2019deep} proposed the VoteNet to vote for the object centroids in a scene. The 3D bounding box is then generated through vote clustering and 3D NMS. Nevertheless, the output of their method is not a complete 6D pose.

Our work is inspired by PVNet~\cite{peng2019pvnet} and VoteNet~\cite{qi2019deep}, and employs deep learning and Hough voting simultaneously to achieve a patch-level 3D Hough voting method for object 6D pose estimation. 

\section{Method}
\label{sec:method}
\subsection{Overview}
\begin{figure*}[!t]
	\begin{center}
		\includegraphics[width=0.95\linewidth]{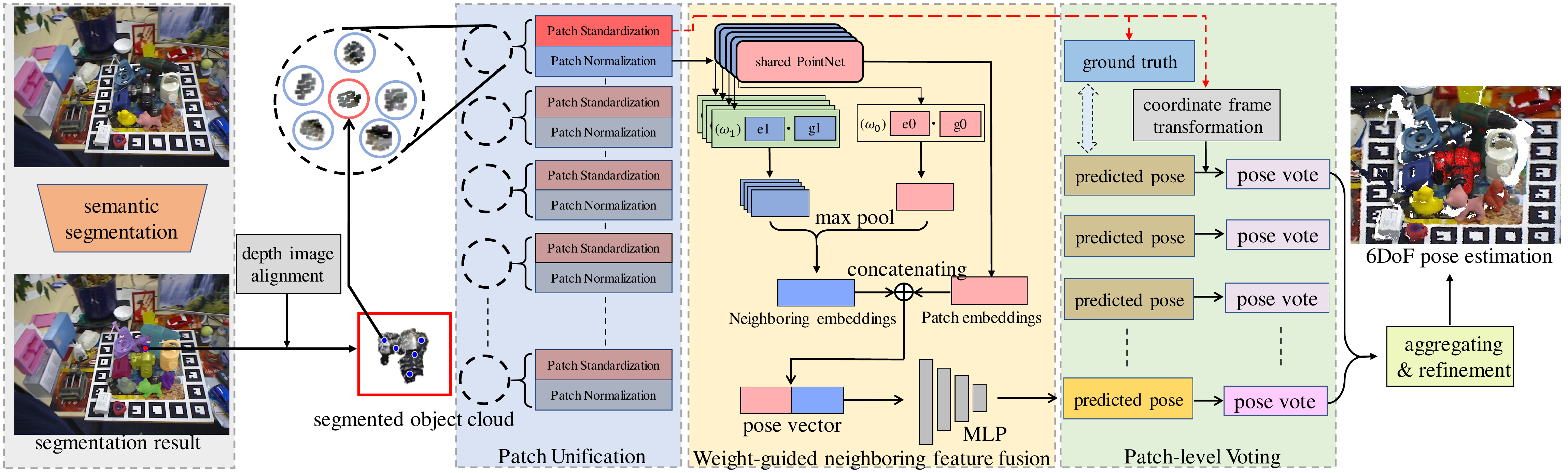}
	\end{center}
	\caption{The architecture of 3DPVNet. The test scene is first segmented to obtain the objects of interest through a semantic segmentation network. A set of patches (the red circle) are then sampled, which are grouped with their neighboring patches (the blue circles). In 3DPVNet, we design three modules: The Patch Unification (PU) module processes the input patch and produces the ground truth and a transformation for pose vote generation; The Weight-guided Neighboring Feature Fusion (WNFF) module fuses features from the center patch and neighboring patches to yield a semi-global feature; The Patch-level Voting (PV) module regresses the transformation vector and generates pose vote. The final object pose is found through a aggregating and refinement step.}
	\label{fig:overview}
\end{figure*}

Given a RGB-D scene, our purpose is to estimate the 6D pose of the objects of interest.
In 3D field, the 6D pose is represented as the 3D translation $t \in \mathbb{R}^3$ and 3D rotation $R \in SO(3)$ of the transformation from the \textit{OCF} to the \textit{GCF}.
In order to mitigate the effects of other classes, including the background, we first utilize a semantic segmentation network to segment the object.
The segmented object is then convert to a point cloud via depth image alignment.
We subsequently propose a Patch-level 3D Hough Voting Network (3DPVNet) to address the pose estimation problem.
Fig.~\ref{fig:overview} summaries the overall architecture.

\subsection{Semantic Segmentation}
\label{subsec:semantic_segmentation}
For a testing scene, not all patches should participate in the voting process. Patches from the background or other objects can interfere with the detection results. To this end, we first filter out the background in the scene and classify the foreground.
Similar to~\cite{wang2019densefusion}, we adopt a semantic segmentation network to achieve this task. 
The network is based on the MaskRCNN~\cite{he2017mask} due to its remarkable performances on image semantic segmentation. We modify the output channels to fit our datasets. The input is an image and the output is a segmentation map. Each pixel in the map is labeled by one of $N+1$ classes, representing $N$ possible classes of objects or the background. We then convert the depth image to point cloud using the intrinsic matrix. Due to the alignment of RGB and depth images, the scene cloud is segmented.

\subsection{6D Pose Estimation with 3DPVNet}
\label{subsec:pose_regression}
3DPVNet takes as input a 3D local patch (center patch) and its $k$ neighboring patches, and outputs a transformation from the local patch coordinate frame to the object coordinate frame, which we called a predicted pose.
We then use the transformation information from the PU module to convert the predicted pose to a pose vote.
After a vote aggregation step and the refinement, the 6D pose is finally determined.

\subsubsection{Network Structure.}
To tackle the three challenges mentioned in the previous section, we present 3DPVNet, which is comprised of three modules (see Fig.~\ref{fig:overview}).
In particular, the \textit{patch unification} (\textbf{PU}) module processes the input $k+1$ 3D patches, and generates normalized patches for network input and a standardized center patch for ground truth generation. The \textit{weight-guided neighboring feature fusion} (\textbf{WNFF}) module extracts representative features for the input patch and its $k$ neighboring patches using a backbone network, and fuses them together to yield a semi-global feature, which is a 2048-D vector. The backbone network we employ is the PointNet architecture~\cite{qi2017pointnet}. The \textit{patch-level voting} (\textbf{PV}) module uses the semi-global feature to regress the local transformation. The regression is implemented through a four-layer MLP, which produces a 6-D transformation vector. The vector is further converted into a pose vote by applying a coordinate frame transformation.

\subsubsection{PU Module.}
\label{subsubsec:training_data}
\begin{figure}[!t]
	\begin{center}
		\includegraphics[width=1\linewidth]{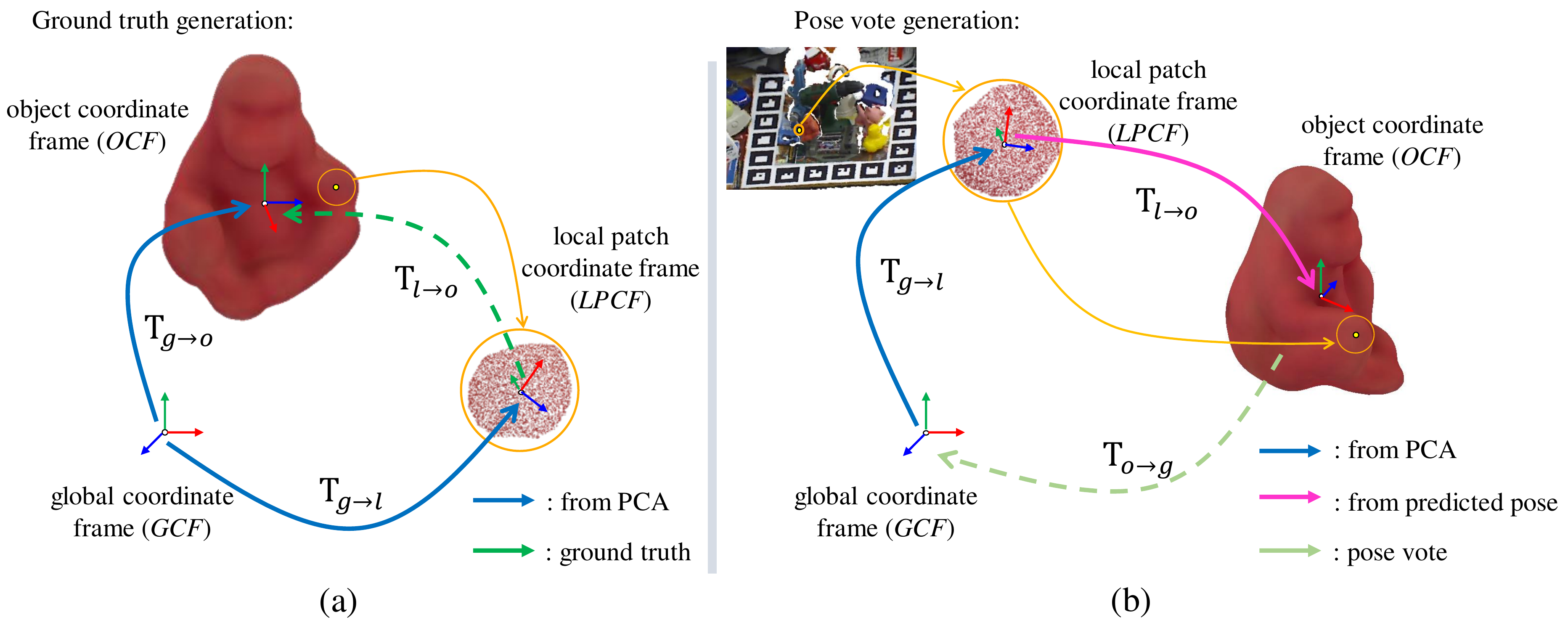}
	\end{center}
	\caption{The processes for ground truth generation and pose vote generation.}
	\label{fig:hough_voting}
\end{figure}

PU module contains two branches: \textit{Patch Normalization} and \textit{Patch Standardization}.
For an input patch $\mathcal{P}_i$, Patch Normalization branch fixes the point number of the patch to $n$ through a randomly up- or down-sampling procedure.
The patch is then de-meaned and normalized for the network input.
On the other hand, Patch Standardization branch builds a standard local coordinate frame on the center patch, namely the \textit{LPCF}, which is invariant to external frames. Patches are originally under the \textit{GCF}, hence the standardized process is formulated as:
\begin{equation}
\label{eq:setlabels}
\hat{\mathcal{P}_0} = \mathbf{T}_{g \rightarrow l}(\mathcal{P}_0),
\end{equation}
where $\mathbf{T}_{g \rightarrow l}(\cdot)$ depicts the standard transformation from the \textit{GCF} to the \textit{LPCF}. $\hat{\mathcal{P}_0}$ is the transformed standard center patch. 
In our work, $\mathbf{T}_{g \rightarrow l}(\cdot)$ is obtained through the Principal Component Analysis (PCA) algorithm~\cite{wold1987principal} due to its stability to slight noise. 
Note, the object model is also standardized as \textit{OCF} before network training in order to keep consistency.
Consequently, the ground truth for the voting is produced by solving
\begin{equation}
\hat{\mathcal{M}} = \mathbf{G}_{l \rightarrow o}(\hat{\mathcal{P}_0}),
\end{equation}
where $\mathbf{G}_{l \rightarrow o}(\cdot)$ represents the transformation from \textit{LPCF} to \textit{OCF}. $\hat{\mathcal{M}}$ is the transformed model. The process is illustrated in Fig.~\ref{fig:hough_voting}(a).
After decomposing the transformation matrix $\mathbf{G}_{l \rightarrow o}(\cdot)$ into a 6-D transformation vector, the ground pose vote is finally obtained.

\subsubsection{WNFF Module.}
For a center patch and $k$ neighboring patches, there are $k+1$ features generated by the backbone network.
Our task is to fuse them together and generate a semi-global feature to describe the local geometric characteristics for the center patch $\mathcal{P}^{(0)}$.
Since the neighboring patches contribute unequally to the description of local geometric characteristics, simply concatenating all the features will magnify the influence of distant patches, which is unfavourable for the pose estimation of occluded objects. 
To remedy this, we propose an adaptive weighting method which can allocate a weight for each patch according to the distance between that patch and the center patch: 
\begin{equation}
\omega_i = 1-dis(c_i, c_0)/D_{obj},
\end{equation}
where $c_0$ and $c_i$ represent the centers of the center patch $\mathcal{P}^{(0)}$ and its $i$-th neighboring patch $\mathcal{P}^{(i)}$, respectively. $dis(\cdot, \cdot)$ is the Euclidean distance function. $D_{obj}$ denotes the diameter of the object, namely the diameter of the Minimum Enclosing Ball of the object model. 

On this basis, in order to make features from different patches have distinguishing effects, we propose to generate a reference vector for each patch after feature learning, with the guidance of its corresponding weight.
The reference vector has the same dimensions with the feature vector (2048-D).
Each dimension is initially set to 0, and has a probability of $\omega$ to turn to 1.
The dot product of the feature vector and the reference vector is then computed, which yields a new feature. After the computation for all $k+1$ feature vectors, we apply max pool operation to the new features to produce a fused feature. 
Our WNFF module can be formulated as:
\begin{equation}
\mathrm{F}^* = \text{maxpool}(g(\omega_0)\cdot\mathrm{F}^{(0)}, g(\omega_1)\cdot\mathrm{F}^{(1)}, \cdots {, g(\omega_k)\cdot\mathrm{F}^{(k)}}), 
\end{equation}
where $\mathrm{F}^*$ is the output feature. $\text{maxpool}(\cdot)$ denotes the max pool operation. $g(\cdot)$ is the reference vector generation function. $\omega_i$ and $k$ represent the weight of $i$-th feature and the number of neighbors, respectively.

\subsubsection{PV Module.}
\label{subsec:hough_voting}
Given a fused feature $\mathrm{F}^*$, we construct a network to regress the 6-D transformation vector. The network is a four-layer MLP. According to the supervision information, the output vector predicts the transformation from the \textit{LPCF} to the \textit{OCF}, i.e., $\widetilde{\mathbf{T}}_{l \rightarrow o}$. 
As is explained above, the object pose $\mathbf{P}$ stands for the rigid transformation of the \textit{OCF} to the \textit{GCF}, which can be marked as $\mathbf{T}_{o \rightarrow g}$.
Hence, the pose vote can be computed through
\begin{equation}
\mathbf{P}^i = \text{inv}(\mathbf{T}_{g \rightarrow l}^i\widetilde{\mathbf{T}}_{l \rightarrow o}^i), 
\end{equation}
where $\mathbf{T}_{g \rightarrow l}$ is the coordinate frame transformation. It is produced by Patch Standardization in the PU module. $\text{inv}(\cdot)$ represents the inverse operation for the transformation matrix. $\mathbf{P}^i$ stands for the pose vote. Moreover, the superscript of $i$ indicates that all transformations are given by the $i$-th center patch. 

In summary, in this module, all votes are cast by local patches, namely the patch-level hough voting. Compared to point-level Hough voting methods, such as PPF~\cite{drost2010model} and PVN3D~\cite{he2020pvn3d}, our method achieves a balance on detection accuracy and computation. Moreover, due to a representative description of the local geometric characteristics, our method is robust to occlusions.

\subsubsection{Votes Aggregating and Refinement.}
Given a set of pose votes $\mathbf{P}^i$ with $i = \{1,2, \cdots, m\}$, where $m$ is the number of center patches from a segmentation, we need to aggregate all votes and output the final object pose, or poses for the case of multiple object instances in the segmentation.
Here, we directly apply the K-means clustering algorithm in pose space to reach the purpose. First, the pose vote $\mathbf{P}$ is decomposed to a 6D pose vector $v$. $K$ seeds are randomly selected over all votes, forming $K$ clusters. We then merge the clusters with the distance in pose space less than a specific threshold $\delta$, and remove the clusters with too few votes. The centers of clusters in pose space are seen as a rough pose estimation result. Finally, we register the object model to the input scene using the rough pose, and utilize an Iterative Closest Point (ICP)~\cite{besl1992method} based refinement to obtain an accurate one.

\subsubsection{Network Training and Testing.}
In the training stage, patches are sampled from 3D models.
More specifically, for an object model $\mathcal{M}$, we apply farthest point sampling method to obtain a sub-sampled point cloud $\mathcal{S}$. Each point in $\mathcal{S}$ generates a local patch $\mathcal{P}$ with points located in a \textit{r}-radius sphere in $\mathcal{M}$. $r$ is setup according to the model diameter, in order to ensure an appropriate coverage to the whole model. $k$ nearest patches are then searched. The training process is supervised by the ground truth from the patch standardization.
In the testing, patches is generated from the segmented object cloud.
Since there is no ground truth, only the patch normalization branch in the PU module takes effect.

\section{Results}
\label{sec:results}
In this section, we primarily focus on the analysis of 3DPVNet in aspects of the robustness to noise, the performance on pose estimation, and the sensibility to heavy occlusions. The experiments are carried out on the UWA dataset~\cite{mian2006three}, the LineMOD dataset~\cite{hinterstoisser2011gradient}, and the bin-picking dataset~\cite{doumanoglou2016recovering}. 
Moreover, we implement the ablation study in terms of the effectiveness of neighboring patches and WNFF module. The influence of the number of neighbors in WNFF module is also studied.

\subsection{Experiment Settings}
\label{subsec:experiment_setting}
\subsubsection{Datasets.}
The datasets we used in the experiments are listed as follows.
\begin{itemize}
	\item UWA dataset. The UWA dataset is a 3D scanned dataset, which contains four complete object models and 50 3D scenes. All four objects are placed manually to cause occlusions and clutter. Each object in each scene is annotated with a ground truth pose, which indicates the transformation from the object model to its real location in the scene. 
	\item LineMOD dataset. The LineMOD dataset contains 15 complete object models. Each object is placed in a cluttered scene to generate test data, which are captured from $\sim$1200 different viewpoints around the scene using a RGB-D sensor. Since our method is mainly based on the point cloud, after the semantic segmentation, we convert all RGB-D scenes in the dataset to 3D point clouds using the given camera intrinsic matrix.
	Note, we train and test our model on 13 objects in the dataset as most methods done in the literature.
	\item Bin-picking dataset. The bin-picking dataset contains two complete object models and 177 RGB-D scenes. In each scene, multiple instances of one or two objects are placed in a small bin, which is very challenging due to the heavy occlusions. The RGB-D scenes are converted to point clouds, as in the LineMOD dataset.
\end{itemize}
All the datasets are acquired through~\cite{hodan2018bop}. For each object model from the datasets, we totally sample nearly 3000 patches from its surface, which are then used to generate the training data and testing data. The ratio of the training data and the testing data is 4:1. 

\subsubsection{Network Setup.}
We set a training epoch of 600, and a batch size of 32. The initial learning rate is 0.001, which is decreased by 10 after 80 epochs, and further decreased by another 10 after 120 epochs. The model is optimized by an Adam optimizer.
The feature embeddings for the local patch is 2048-D.
The size of layers in the four-layer MLP of the pose regression network is of 1024, 512, 256, 6, respectively.
The number of neighbors of a patch is 8 in all experiments, which will be further analyzed.
The output tensor of the semantic network is set to 4+1 for the UWA dataset, 15+1 for the LineMOD dataset, and 2+1 for the bin picking dataset, where ``1'' is for the background class.
The network is trained on a PC with 16GB memory, 12 Intel(R) Core(TM) i7-8700K CPU, and a NVIDIA GeForce GTX 1080 GPU.

\subsubsection{Metrics.}
We use two metrics in our experiments. 
For the UWA dataset and the bin-picking dataset, we use F1 score~\cite{hinterstoisser2011gradient} to evaluate the performance of methods. F1 score is the harmonic mean of precision and recall, which serves as an accurate form of comparison.
For the LineMOD dataset, we use the ADD~\cite{hinterstoisser2012model} for non-symmetric objects and ADD-S for symmetric objects. ADD(-S) measures the average distance of all model points transformed using the ground truth pose and the estimated pose. The chosen coefficient in the metric is set to 0.1 for the experiments.

\subsection{Evaluations on the UWA Dataset}
\label{subsec:evaluation_uwa}
We first test our method on the UWA dataset.
For comparative results, we compare the performance of our method against Mian \textit{et al.}~\cite{mian2006three}, PPF~\cite{drost2010model}, and one of its variations~\cite{hinterstoisser2016going} (refer to PPF+ in the following), latent-class Hough Forest~\cite{tejani2014latent} (refer to LCHF in the following), and Buch \textit{et al.}~\cite{buch2017rotational}.
The results are shown in Tab.~\ref{tab:qresult_uwa_f1}. It is obvious that our method reaches the best over all testing methods. 
PPF~\cite{drost2010model} has difficulty on dealing with scenes with heavy noise, and thus shows the poorest performance. By contrast, 3DPVNet employs the PU module to construct a standard form for the input patch, which is tolerant to a certain level of noise. 

\begin{table}
	\small
	\def\arraystretch{1.2}
	\centering
	\caption{F1 scores of Mian \textit{et al.}~\cite{mian2006three}, PPF~\cite{drost2010model}, PPF+~\cite{hinterstoisser2016going}, LCHF~\cite{tejani2014latent}, Buch \textit{et al.}~\cite{buch2017rotational} and 3DPVNet on the UWA dataset.}
	\label{tab:qresult_uwa_f1}
	{
		\begin{tabular}{l|c}
			\hline
			\textbf{Approaches} & \textbf{F1 Score}  \\
			\hline
			\hline
			\multirow{1}{*}{Mian~\cite{mian2006three}} & 0.56 \\
			\hline
			\multirow{1}{*}{PPF~\cite{drost2010model}} & 0.62 \\
			\hline
			\multirow{1}{*}{PPF+~\cite{hinterstoisser2016going}} & 0.84 \\
			\hline
			\multirow{1}{*}{LCHF~\cite{tejani2014latent}} & 0.51 \\
			\hline
			\multirow{1}{*}{Buch~\cite{buch2017rotational}} & 0.87 \\
			\hline
			\multirow{1}{*}{3DPVNet (Ours)} & \textbf{0.93} \\
			\hline
		\end{tabular}
	}
\end{table}

\begin{table}
	\scriptsize
	\def\arraystretch{1.2}
	\centering
	\caption{ADD of PointFusion~\cite{xu2018pointfusion}, Sundermeyer \textit{et al.}~\cite{sundermeyer2018implicit}, Kehl \textit{et al.}~\cite{kehl2017ssd}, DeepIM~\cite{li2018deepim}, DenseFusion~\cite{wang2019densefusion} and our method on the LineMOD dataset. Objects with italic name are symmetric.}
	\label{tab:qresult_lmd_ms}
	{
		\begin{tabular}{l|p{1.1cm}<{\centering}|p{0.6cm}<{\centering}|p{0.4cm}<{\centering}|p{0.65cm}<{\centering}|p{1.1cm}<{\centering}|p{0.5cm}<{\centering}}
			\hline
			\textbf{Approaches} & PointFusion & Implicit & Kehl & DeepIM & DenseFusion & Ours  \\
			\hline
			\hline
			\multirow{1}{*}{Sequence} & \multicolumn{6}{c}{ADD} \\
			\hline
			\multirow{1}{*}{Ape} & 0.704 & 0.206 & 0.65 & 0.77 & 0.923 & \textbf{0.952} \\
			\hline
			\multirow{1}{*}{Bench Vise} & 0.807 & 0.643 & 0.80 & \textbf{0.975} & 0.932 & 0.947 \\
			\hline
			\multirow{1}{*}{Camera} & 0.608 & 0.58 & 0.632 & 0.935 & 0.944 & \textbf{0.959} \\
			\hline
			\multirow{1}{*}{Can} & 0.611 & 0.761 & 0.86 & \textbf{0.965} & 0.931 & 0.943 \\
			\hline
			\multirow{1}{*}{Cat} & 0.791 & 0.720 & 0.70 & 0.821 & \textbf{0.965} & 0.940 \\
			\hline
			\multirow{1}{*}{Driller} & 0.473 & 0.416 & 0.73 & 0.950 & 0.870 & \textbf{0.973} \\
			\hline
			\multirow{1}{*}{Duck} & 0.630 & 0.325 & 0.66 & 0.777 & 0.923 & \textbf{0.933} \\
			\hline
			\multirow{1}{*}{\textit{Egg box}} & 0.999 & 0.986 & 1.00 & 0.971 & \textbf{0.998} & 0.985 \\
			\hline
			\multirow{1}{*}{\textit{Glue}} & 0.993 & 0.964 & 1.00 & 0.994 & \textbf{1.000} & 0.982 \\
			\hline
			\multirow{1}{*}{Hole Punch} & 0.718 & 0.58 & 0.499 & 0.528 & 0.921 & \textbf{0.938} \\
			\hline
			\multirow{1}{*}{Iron} & 0.832 & 0.631 & 0.78 & \textbf{0.983} & 0.970 & 0.962 \\
			\hline
			\multirow{1}{*}{Lamp} & 0.623 & 0.917 & 0.73 & 0.975 & 0.953 & \textbf{0.981} \\
			\hline
			\multirow{1}{*}{Phone} & 0.788 & 0.710 & 0.79 & 0.877 & \textbf{0.928} & 0.922 \\
			\hline
			\hline
			\multirow{1}{*}{\textbf{Average}} & 0.737 & 0.647 & 0.79 & 0.886 & 0.943 & \textbf{0.955} \\
			\hline
		\end{tabular}
	}
\end{table}

\begin{figure*}[!t]
	\begin{center}
		\includegraphics[width=0.85\linewidth]{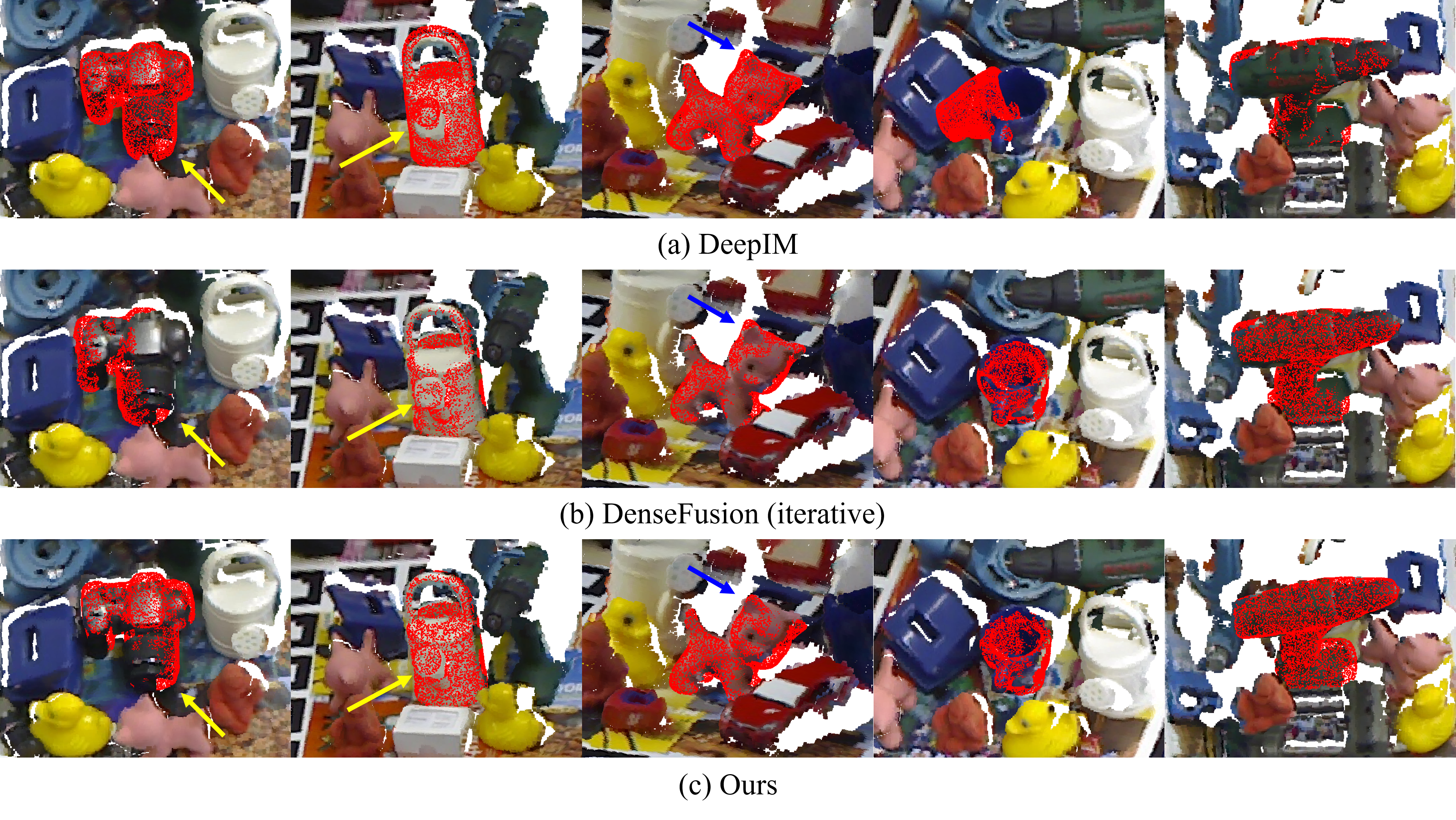}
	\end{center}
	\caption{Visualized results of our method (c) against DeepIM~\cite{li2018deepim} (a) and DenseFusion~\cite{wang2019densefusion} (b) on the LineMOD dataset.}
	\label{fig:vresult_lmd}
\end{figure*}

\subsection{Evaluations on the LineMOD Dataset}
\label{subsec:evaluation_lmd}
We then test our method on the LineMOD dataset.
Rows 1-3 of Fig.~\ref{fig:vresult_lmd} show the visualized results. As can be seen, our method achieves satisfactory results, despite the challenge of the cluttered background in the testing scenes. Furthermore, in the third row, we add an experiment for a symmetric object (the glue in the LineMOD dataset). The result demonstrates the effectiveness of our method for recognizing symmetric objects.
\subsubsection{Quantitative Evaluation.}
Comparative results are displayed in Tab.~\ref{tab:qresult_lmd_ms}. Here, we compare our method against LineMOD~\cite{hinterstoisser2011gradient}, PPF+~\cite{hinterstoisser2016going}, Kehl \textit{et al.}~\cite{kehl2017ssd}, Buch \textit{et al.}~\cite{buch2017rotational}, DeepIM~\cite{li2018deepim}, DenseFusion \textit{et al.}~\cite{wang2019densefusion}. As can be seen, our method outperforms the existing methods for most objects in the dataset. Note that for the glue and the egg box, the recent DenseFusion performs better than ours. The main reason is that our method votes for the object pose using a local region, which has a relatively weak ability to deal with the symmetric objects. 

We compare 3DPVNet with PointFusion~\cite{xu2018pointfusion}, Sundermeyer \textit{et al.}~\cite{sundermeyer2018implicit}, Kehl \textit{et al.}~\cite{kehl2017ssd}, DeepIM~\cite{li2018deepim}, and DenseFusion~\cite{wang2019densefusion}. The evaluation results are reported in Table.~\ref{tab:qresult_lmd_ms}. Note that the performances of other methods are from the literature. From the results, we can see that our method outperforms DenseFusion by 1.2\%, and achieves a large improvement to DeepIM by 6.9\%. It can be inferred that RGB-only methods such as DeepIM cannot extract geometric characteristics, and therefore are difficult to improve their performances into a relatively high level. Comparatively, our method directly takes the 3D local patch as input, which has a better description to local geometric characteristics. Moreover, the designed eight-guided feature fusion module also has the effectiveness to mining contextual information around the patch.

\subsubsection{Qualitative Evaluation.}
Fig.~\ref{fig:vresult_lmd} displays the detection results by DeepIM~\cite{li2018deepim}, iterative DenseFusion~\cite{wang2019densefusion} and our method. Obviously, our method performs the best over all methods. For DeepIM, the matching results are not very good in most scenes, especially in the fourth column, which is a challenging occluded scene. One of the reasons is that DeepIM utilizes the object image and mask to regress the pose, which is less robust to occlusions. Furthermore, the cluttered environment also disturbs the estimation. In terms of DenseFusion, the estimation results are not highly accurate in the first three scenes. The reason may be that the objects in these scenes are extremely incomplete, causing a decent capability on describing the global feature from 3D point cloud. By contrast, our method adopts a patch-level voting scheme, which is more stable on extracting local geometric characteristics, and is also robust to occlusions.

\subsection{Evaluations on the Bin-picking Dataset}
\label{subsec:evaluation_bpg}
We further evaluate 3DPVNet on the bin-picking dataset. 
Tab.~\ref{tab:qresult_bpg_f1} reports the F1 score of our method against Doumanoglou \textit{et al.}~\cite{doumanoglou2016recovering}, LCHF~\cite{tejani2014latent}, Buch \textit{et al.}~\cite{buch2017rotational} on the bin-picking dataset. On average, our method is superior to the existing methods. 
The reason can be concluded that 3DPVNet takes into consideration the local geometric characteristics through the WNFF module, resulting in a better capability to tackle heavy occlusions. 

\subsection{Ablation Study}
\label{subsec:ablation_study}
\begin{figure}[!t]
	\begin{center}
		\includegraphics[width=0.9\linewidth]{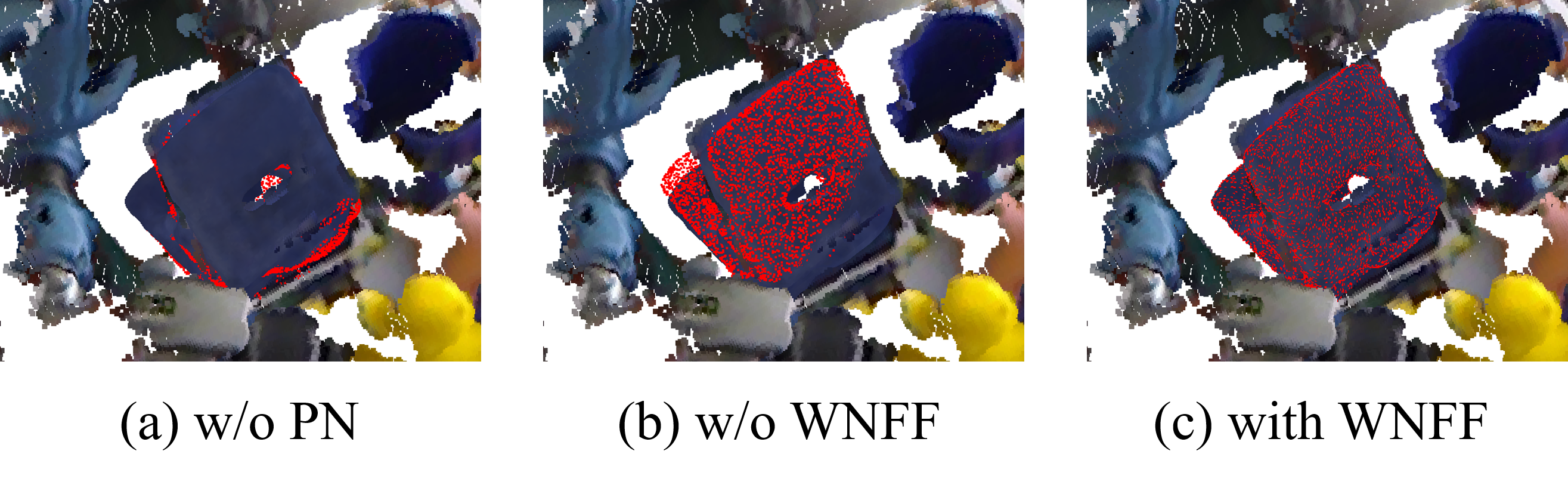}
	\end{center}
	\caption{Visualized results for ablation study. (a) without neighboring patches. (b) without weight-guided neighboring feature fusion module. (c) the proposed method with weight-guided neighboring feature fusion module equipped.}
	\label{fig:ablation_study}
\end{figure}

\begin{table}
	\def\arraystretch{1.2}
	\centering
	\caption{Ablation study of 3DPVNet.}
	\label{tab:ablation_study}
	{
		\begin{tabular}{c|cc|c}
			\hline
			\textbf{Method} & \textbf{NP} &  \textbf{WNFF} & \textbf{ADD} \\
			\hline
			3DPVNet (Ours) &  &  & 0.634 \\
			\hline
			3DPVNet (Ours) &  $\surd$ &   & 0.846 \\
			\hline
			3DPVNet (Ours) &  $\surd$ &  $\surd$ & \textbf{0.955} \\
			\hline
		\end{tabular}
	}
\end{table}

To experimentally evaluate the performances of different choices of our method, we conduct several ablation experiments. Some notations are listed:
\begin{itemize}
	\item[$\bullet$] \textbf{w/o neighboring patches (NP)}. The network is trained and tested only with the reference patch.
	\item[$\bullet$] \textbf{w/o weight-guided neighboring feature fusion module (WNNF)}. Without the weight-guided neighboring feature fusion module, the features from neighboring patches will directly concatenate together.
\end{itemize}
The results are displayed in Table.~\ref{tab:ablation_study}. As we can see from the second and the third row, when adding the neighboring patches to our network, the performance is significantly improved from 0.634 to 0.816, which proves that adding neighboring patches can enhance the representation capability of geometric characteristics. By adding the novel weight-guided neighboring feature fusion module, the performance is improved to 0.937, which demonstrates the effectiveness of the proposed module.	
The visualized results are shown in Fig.~\ref{fig:ablation_study}. Obviously, with the weight-guided neighboring feature fusion module, the best performance is achieved.

\begin{table}
	\footnotesize
	\def\arraystretch{1.2}
	\centering
	\caption{F1 scores of Doumanoglou \textit{et al.}~\cite{doumanoglou2016recovering}, LCHF~\cite{tejani2014latent}, Buch \textit{et al.}~\cite{buch2017rotational} and our method on the bin-picking dataset.}
	\label{tab:qresult_bpg_f1}
	{
		\begin{tabular}{l|c|c|c|c}
			\hline
			\textbf{Method} & Doumanoglou & LCHF & Buch & Ours \\
			\hline
			\hline
			\multirow{1}{*}{Sequence} & \multicolumn{4}{c}{F1 score} \\
			\hline
			\multirow{1}{*}{Juice Carton} & 0.62 & 0.49 & 0.81 & \textbf{0.87} \\
			\hline
			\multirow{1}{*}{Coffee Cup} & 0.53 & 0.54 & 0.76 & \textbf{0.84} \\
			\hline
			\hline
			\multirow{1}{*}{\textbf{Average}} & 0.58 & 0.52 & 0.79 & \textbf{0.86} \\
			\hline
		\end{tabular}
	}
\end{table}

\subsubsection{The study of the Number of Neighbors.}
The number of neighbors of a patch, referred as $k$ in previous section, has a significant impact on the representation capability of local geometric characteristics. 
A higher value leads to a representative description, since the ambiguities due to similar regions from the object can be reduced. Nevertheless, the higher value tends to be less robust to incomplete data in the testing stage, as the neighboring patches may not be the same as those in the training stage during which the neighboring patches arise from a complete model.
Therefore, a proper value with consideration of descriptiveness and robustness should be investigated. To this end, we conduct a series of experiments to evaluate the impact of $k$, which guides us to choose a reasonable one.

\begin{figure}[!t]
	\begin{center}
		\includegraphics[width=0.85\linewidth]{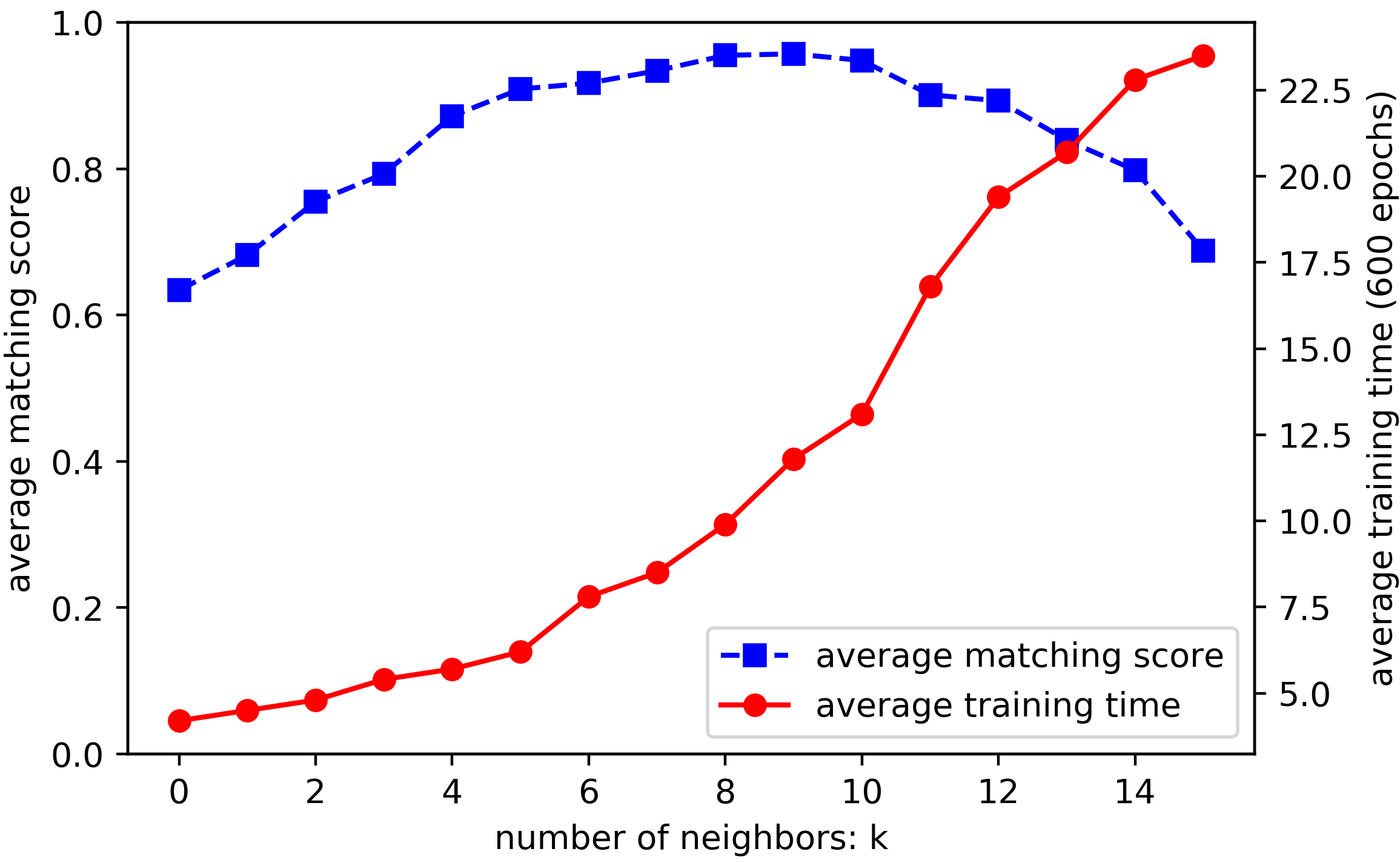}
	\end{center}
	\caption{The curves for ADD (the blue dot line) and average training time (hours for 600 epochs, the red line) along different settings of $k$ in the dataset of LineMOD.}
	\label{fig:evaluation_of_k}
\end{figure}

The experiments are implemented by increasing $k$ from 0 to 15 in steps of 1. For each $k$, we record the average training time (in hour) of 600 epochs and compute the ADD over 13 objects in the LineMOD dataset. Note, for $k=0$, we do not use any neighboring patch and the proposed weight-guided neighboring feature fusion module, which is the same as the case of \textbf{w/o NP} in the ablation study. The results are illustrated in Fig.~\ref{fig:evaluation_of_k}.
As can be seen, with the increasing of $k$, the average training time (displayed in red line) is rising up rapidly. The reason is that the amount of learnable parameters in the network is exploding, leading to a higher training time. In terms of the ADD (represented in blue dot line), the tendency is similar to our analysis above. We observe that satisfactory results can be achieved when $k$ is around 8. To get a balance with training time, we set $k$ to 8 in all experiments, as introduced in the network setup above.

\section{Conclusion}
In this paper, we introduce 3DPVNet for 6D pose estimation in point cloud. Our method is a patch-level Hough voting method that based on pose regression. The patches are regressed locally to produce 6D pose votes, which are then cast to vote for the final object pose. Since the pose regression model is built locally, our method is significantly robust to occlusions. We also design a weight-guided neighboring feature fusion module in the pose regression network which utilizes the novel guided tensor to fuse features from the neighboring patches and the reference patch, making it possible to enhance geometric characteristics description. Furthermore, the patch-level Hough voting achieves a less computation than point/pixel-level voting methods. A series of experiments on the LineMOD dataset, including quantitative and qualitative evaluations, demonstrate that our method outperforms other methods.

{\small
\bibliographystyle{ieee}
\bibliography{egbib}
}

\end{document}